\documentclass{ieeetj}
\usepackage{cite}
\usepackage{amsmath,amssymb,amsfonts}
\usepackage{algorithmic}
\usepackage{graphicx,color}
\usepackage{textcomp}
\usepackage{xcolor}
\usepackage{hyperref}
\hypersetup{hidelinks=true}
\usepackage{algorithm,algorithmic}
\def\BibTeX{{\rm B\kern-.05em{\sc i\kern-.025em b}\kern-.08em
    T\kern-.1667em\lower.7ex\hbox{E}\kern-.125emX}}
\AtBeginDocument{\definecolor{tmlcncolor}{cmyk}{0.93,0.59,0.15,0.02}\definecolor{NavyBlue}{RGB}{0,86,125}}

\usepackage{array}
\usepackage{booktabs}
\usepackage[table]{xcolor}

\def\OJlogo{\vspace{-4pt}Hallucination in Medical AI}
\def\seclogo{\vspace{10pt}Hallucination in Medical AI}

\def\authorrefmark#1{\ensuremath{^{\textbf{#1}}}}

\begin{document}
\receiveddate{XX Month, XXXX}
\reviseddate{XX Month, XXXX}
\accepteddate{XX Month, XXXX}
\publisheddate{XX Month, XXXX}
\currentdate{XX Month, XXXX}
\doiinfo{XXXX.2022.1234567}

\markboth
{Hallucination in Medical Imaging AI}
{Omar Alshahrani and Muzammil Behzad}

\title{Hallucination in Medical Imaging AI: A Cross-Modality Analytical Framework for Taxonomy, Detection, and Mitigation under Regulatory Constraints}

\author{\uppercase{Omar Alshahrani}\authorrefmark{1} and
\uppercase{Muzammil Behzad}\authorrefmark{1, 2}}

\affil{King Fahd University of Petroleum \& Minerals, Saudi Arabia}
\affil{SDAIA-KFUPM Joint Research Center for Artificial Intelligence, Saudi Arabia}

\corresp{Corresponding author: Muzammil Behzad (e-mail: muzammil.behzad@kfupm.edu.sa).}

\authornote{This work was funded by King Fahd University of Petroleum and Minerals, Saudi Arabia. The authors would also like to acknowledge the computational support from SDAIA-KFUPM Joint Research Center for Artificial Intelligence, Saudi Arabia.}

\begin{abstract}
AI systems are being deployed across medical imaging faster than their failure modes
are understood. At this point in time, the failure of greatest clinical concern is hallucination: clinically
plausible but factually incorrect outputs, including fabricated anatomical structures,
missed findings, incorrect laterality, and invented measurements in generated reports,
with direct consequences, for example, for biopsy decisions, staging, and treatment planning. This
structured narrative synthesizes peer-reviewed studies, benchmark datasets, and
FDA regulatory guidance (2023-2026) across five imaging modalities, such as CT, MRI,
PET/SPECT, ultrasound, and digital pathology, to produce a cross-modality analysis
of hallucination taxonomy, etiology, detection, and mitigation. Specifically, we address
three questions in this study: (1) how can existing taxonomies be unified across modalities?, (2) how do
medical-specialized foundation models hallucinate less than general-purpose ones?, and
(3) which mitigation strategies are effective and compatible with FDA lifecycle
oversight? We note that three taxonomic frameworks together cover the imaging pipeline
in a way no single framework does alone. We also highlight that general-purpose
foundation models outperform medical-specialized models on hallucination-specific
benchmarks (with median hallucination-free rate 76.6\% vs.\ 51.3\%; $p=0.012$), indicating
that narrow domain fine-tuning can introduce overfitting-induced confabulation. At the
same time, the oversight of radiologists remains essential; for instance, a very high percentage of of AI-generated flags
required expert correction before clinical use. Physics-informed architectural constraints,
Chain-of-Thought prompting (with up to 86.4\% hallucination reduction), and human-in-the-loop
safeguards each address different failure modes and is effective when combined. All findings are
mapped to the FDA's Total Product Lifecycle (TPLC) and Predetermined Change Control Plan
(PCCP) frameworks, which treat hallucination management as a lifecycle obligation rather
than a pre-deployment checklist.
\end{abstract}



\maketitle

\section{Introduction}
\label{sec:introduction}

\PARstart{I}{n} diagnostic imaging, model errors can have direct and significant clinical consequences, particularly when they influence critical decision-making processes.
A missed pulmonary nodule, an incorrectly lateralized finding, or a fabricated uptake
pattern on a PET scan can redirect a patient toward inappropriate treatment pathways.
Vision-language models are now being deployed to interpret radiographs, MRI volumes, CT
scans, and digital pathology slides, and their rapid adoption has outpaced the development
of reliable methods for evaluating how these systems fail \cite{rajpurkar2022}.

The failure mode of greatest clinical concern is hallucination, which has emerged as a
critical limitation in the deployment of AI systems in medical imaging. Following Kim
et al.\ \cite{kim2025} and the MediHall framework \cite{chen2024}, we define a
hallucination as a model output that is linguistically or visually coherent but not
supported by the input image or the patient's clinical context. In medical imaging,
this includes fabricated anatomical structures, misclassified pathologies, invented
measurement values in generated reports, and findings that are absent from the underlying
image. These are not merely cosmetic artifacts; in high-acuity clinical settings, such
errors can propagate through workflows and reach patients before they are detected
\cite{kim2025,fda2024tplc}.

The research and regulatory communities have responded to these challenges on multiple fronts, reflecting the growing urgency of addressing hallucination in medical AI systems. Taxonomic
frameworks have been proposed, some of which are modality-specific, while others are
based on severity grading. Benchmark efforts include Med-HallMark (2024), which
provides over 20,000 QA pairs for systematic evaluation~\cite{medhallbench2024}, and CXR-VisHal, which extends
assessment to visual grounding in chest radiographs~\cite{medhallbench2024}. The FDA has moved in parallel,
with the January 2025 draft guidance on AI-enabled device software functions
\cite{fda2025draft} and the December 2024 PCCP finalization \cite{fda2024pccp}
collectively reframing hallucination management as a continuous regulatory obligation
rather than a pre-deployment checklist.

Three key gaps remain unresolved in the current literature. First, existing taxonomies
have been developed for individual imaging modalities and have not been validated
across modalities; consequently, no unified framework currently spans CT, MRI,
PET/SPECT, ultrasound, and digital pathology. Second, the hallucination behavior of
general-purpose versus medical-specialized foundation models has not been directly
compared using hallucination-specific benchmarks, despite important implications for
procurement and deployment decisions. Third, mitigation strategies have largely been
evaluated in controlled benchmark settings, with limited evidence of their effectiveness
in real-world clinical workflows.

This research aims to achieve four primary objectives: (1) to classify hallucination
types into a unified cross-modality taxonomy, (2) to compare general-purpose and
medical-specialized foundation models using hallucination-specific benchmarks,
(3) to evaluate detection and mitigation methods across the five imaging modalities
described above, and (4) to align the resulting evidence with the FDA's TPLC and PCCP
frameworks. A structured narrative methodology was selected over strict meta-analytic
approaches because the existing literature is too heterogeneous in study design,
benchmark selection, and outcome measures to support a formal meta-analysis.

The rest of the paper is organized as follows. Section~\ref{sec:background}
situates the work within prior taxonomic and benchmarking efforts.
Section~\ref{sec:methodology} details the methodology.
Section~\ref{sec:results} presents the results and discussion.
Section~\ref{sec:conclusions} concludes the paper and identifies priority
directions for future research.

\section{Background and Related Work}
\label{sec:background}

\subsection{Clinical Impact of Hallucination in AI-Assisted Diagnostics}

Vision-language models are increasingly used to interpret radiographs, CT scans, MRI volumes, PET/SPECT
acquisitions, and digital pathology slides; however, their outputs are not always reliably correct. The central
failure mode is hallucination, defined as output that is clinically coherent but factually incorrect. Documented
error types include fabricated anatomical structures that are not present in the image \cite{chen2024},
mischaracterized pathologies (for example, GPT-4V has been shown to assert pleural effusions on normal chest
radiographs \cite{kim2025}), incorrect laterality or measurement values embedded in generated reports, and
the omission of clinically significant findings. Prevalence estimates vary by task and model: on Med-HallMark,
general-purpose vision-language models (VLMs) produced hallucinated content in a substantial fraction of cases
\cite{medhallbench2024}, and CXR-VisHal reported grounding failures on chest radiograph questions at
comparable rates \cite{medhallbench2024}. These errors are not merely abstractions of data quality; in radiology
and nuclear medicine, AI outputs feed directly into biopsy decisions, staging classifications, and treatment plans.
Consequently, any undetected hallucination represents a potential source of patient-level harm
\cite{kim2025,rajpurkar2022}.

\subsection{Taxonomies, Benchmarks, and Regulatory Perspectives}
The research on hallucination originated in natural language processing (NLP), primarily in
text-only tasks such as summarization and question answering. The resulting taxonomies
typically divide errors into intrinsic hallucinations, which contradict the source,
and extrinsic hallucinations, which introduce content not present in the source
\cite{ji2023}. However, when vision-language models were applied to the medical domain,
these two categories proved too coarse, as spatial relationships, anatomical constraints,
and imaging physics are not adequately captured by the intrinsic/extrinsic distinction.

Three domain-specific frameworks are most frequently cited in the medical imaging
literature, each categorizing hallucinations along a distinct axis. The Brooks and
Anastasio taxonomy \cite{brooks2025} organizes machine hallucinations in radiology
according to their etiological origin, distinguishing errors attributable to corrupt
or unrepresentative training data from those arising from model confabulation, defined
as outputs with no referent in the input image. This etiological axis was designed to
support targeted remediation by clarifying whether an observed hallucination requires
a data-level intervention, a model-level adjustment, or both. The DREAM Report taxonomy was developed for nuclear medicine and is structured around
clinical risk categories specific to PET/SPECT reporting, including fabricated uptake
patterns, incorrect anatomic localization of tracer activity, and quantitative errors
in SUV reporting \cite{xia2025}. The MediHall hierarchy \cite{chen2024} targets Med-VQA
and large vision-language models (LVLMs) and organizes hallucinations according to
clinical severity within a five-tier framework, ranging from minor errors that do not
affect diagnosis to catastrophic errors that would alter clinical management.

Benchmarks have been developed alongside these taxonomies to support systematic
evaluation of hallucination in medical imaging models. Med-HallMark (2024) provides
over 20,000 QA pairs for measuring hallucination across Med-VQA tasks. MedHallBench
extends evaluation to report generation by assessing whether generated radiology
reports contain findings that are absent from the image. CXR-VisHal further expands
evaluation to chest radiograph visual grounding by testing whether models can localize,
in addition to identifying, a given finding. MIMIC-CXR-VQA and the broader MIMIC-CXR
benchmark family remain standard reference datasets for chest imaging
\cite{medhallbench2024}.

The FDA's regulatory position has evolved in parallel with these developments. The
January 2025 draft guidance \cite{fda2025draft} and the December 2024 PCCP
finalization \cite{fda2024pccp} together establish hallucination management as a
post-market, lifecycle obligation rather than a pre-deployment checkpoint.

\subsection{Open Challenges and Research Gaps}

Three critical gaps remain insufficiently addressed in the current literature on
hallucination in medical imaging, particularly in relation to cross-modality
generalization, comparative model behavior, and real-world clinical validation.

\begin{enumerate}
    \item \textbf{Cross-modality taxonomic validation.} Existing taxonomies have been
developed for specific imaging modalities and have not been validated across CT, MRI,
PET/SPECT, ultrasound, and digital pathology simultaneously, leaving practitioners
without a shared vocabulary for comparing hallucination types across imaging contexts.

    \item  \textbf{General vs.\ specialised model comparison.} The hallucination behavior
of general-purpose foundation models versus medical-specialised foundation models has
not been directly compared using hallucination-specific benchmarks. As a result,
procurement decisions are currently based on an untested assumption, namely that
specialisation reduces hallucination, which has not been evaluated in the
hallucination-specific literature.

    \item \textbf{Real-world and regulatory evidence.} Mitigation strategies have largely
been evaluated in controlled benchmark settings, with limited evidence from deployed
clinical workflows. Furthermore, alignment between reported mitigation performance
and the FDA's TPLC and PCCP expectations has not been systematically examined.
\end{enumerate}

This work is therefore structured around these three gaps, with the aim of providing
a unified analytical perspective that integrates taxonomy, empirical evaluation, and
regulatory alignment across medical imaging modalities.

\section{Framework for Cross-Modality Hallucination Assessment}
\label{sec:methodology}

\subsection{Research Design and Analytical Framework}

This study adopts a structured narrative design, drawing on peer-reviewed empirical studies,
regulatory guidance documents, and clinical evaluation frameworks. A strict systematic meta-analysis protocol was not feasible, as the
source literature spans algorithmic AI research, clinical radiology, nuclear
medicine, digital pathology, and regulatory science, with heterogeneous metrics
that cannot be directly pooled. The work is organized around four analytical
areas: taxonomic classification, etiological analysis, evaluation of detection methodologies,
and assessment of mitigation strategies.

\subsection{Literature Scope and Source Selection}

\begin{figure}[t!]
  \centering
  \includegraphics[width=\columnwidth]{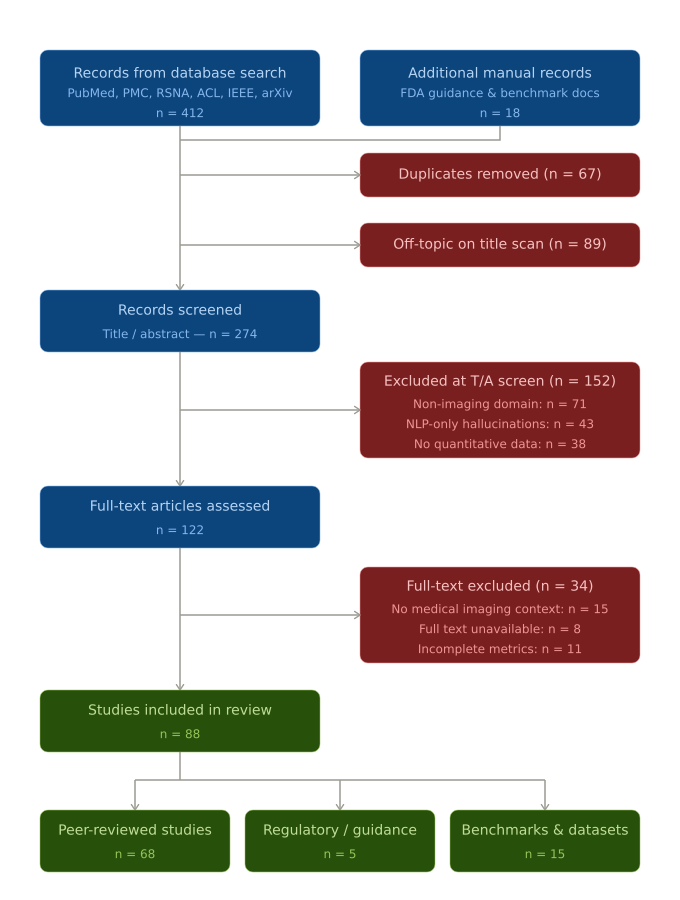}
  \caption{Study selection flow diagram for our structured narrative, illustrating record identification from PubMed, PMC, RSNA Journals, ACL Anthology, IEEE Xplore, and arXiv, as well as targeted retrieval of FDA regulatory guidance and benchmark dataset documentation. Screening was conducted through title/abstract and full-text stages using pre-specified inclusion and exclusion criteria. (Authors' visualization based on data from the sources listed above.)}
  \label{fig:flow}
\end{figure}

As shown in Fig. \ref{fig:flow},
the sources considered in this study fall into three categories. Peer-reviewed
publications were drawn from PubMed, RSNA Journals, ACL Anthology, arXiv, and PMC,
with a focus on studies involving CT, MRI, PET/SPECT, radiography, ultrasound, and
digital pathology. Regulatory documents were obtained from the FDA, including the
January 2025 draft guidance \cite{fda2025draft} and the December 2024 PCCP
finalization \cite{fda2024pccp}. Benchmark and dataset documentation included
Med-HallMark (20,000+ QA pairs across MIMIC-test and OpenI), MedHallBench, ROCO,
Path-VQA, and the FDA-supported sFRC toolkit.
\begin{figure*}[t!]
  \centering
  \includegraphics[width=\textwidth]{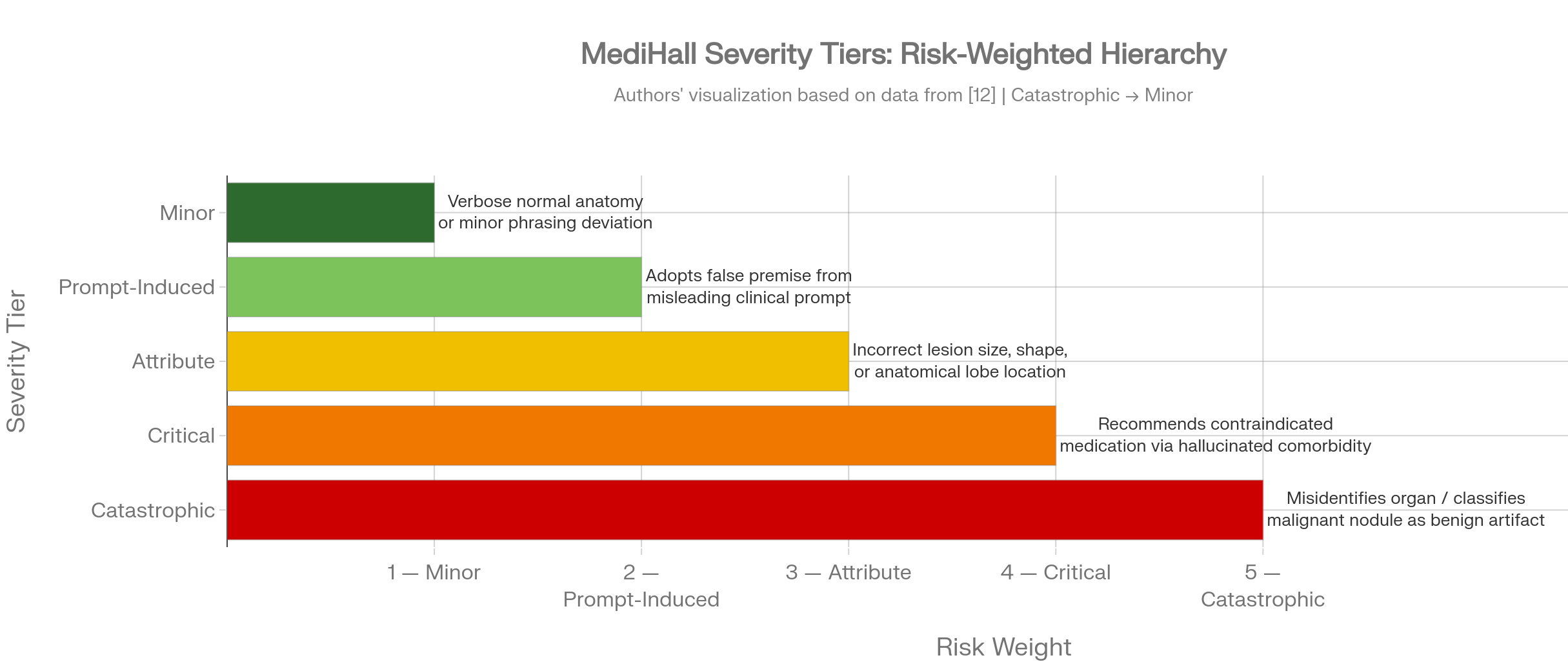}
  \caption{MediHall five-tier hallucination severity hierarchy, a risk-weighted classification ranging from catastrophic to minor. (Authors' visualization based on data from \cite{chen2024}.)}
  \label{fig:medihall}
\end{figure*}

We included those studies which addressed one or more of the following: formal
hallucination taxonomies in a medical imaging context; empirical quantification of
hallucination rates or false-positive burdens; detection frameworks applied to vision
or vision-language models; or architectural and workflow-level mitigation
interventions with reported performance metrics. Studies focused exclusively on
non-clinical, general-domain NLP hallucinations without relevance to medical imaging
were excluded.

\subsection{Taxonomic Classification Framework}

To organize hallucination types across the literature, this work
cross-references three domain-specific taxonomic frameworks. The first is the framework proposed by Brooks and Anastasio \cite{brooks2025}, which
organizes machine hallucinations in radiology along an etiological axis that
distinguishes data-driven failures from model confabulation. The second is the DREAM
Report framework for nuclear medicine imaging \cite{xia2025}, which distinguishes
between factual hallucinations, defined as contradictions of anatomical truth, and
faithfulness hallucinations, defined as deviations from input-specific spatial context.
These distinctions are particularly critical in PET/SPECT modalities, where
AI-generated content may be misinterpreted as metastatic uptake. As demonstrated in Fig.~\ref{fig:medihall}, the third is the
MediHall hierarchical categorization~\cite{chen2024}, developed for LVLMs in Med-VQA
and imaging report generation tasks, which stratifies hallucinations into five
clinically graded severity tiers: catastrophic, serious, moderate, minor-significant,
and minor.

\subsection{Etiological Analysis}

\begin{figure*}[htbp]
  \centering
  \includegraphics[width=\textwidth]{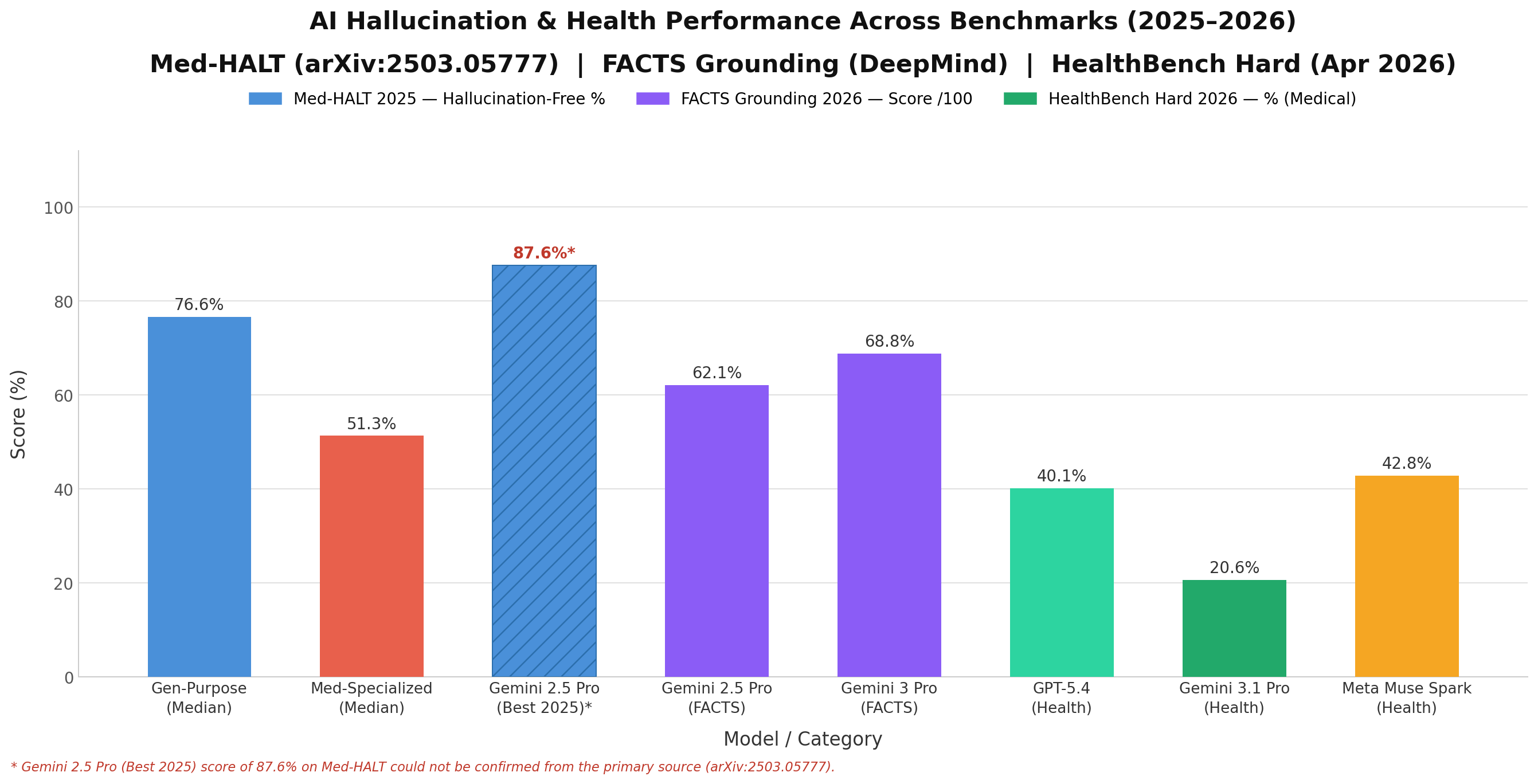}
  \caption{Hallucination and health benchmark performance across frontier models (2025--2026). Left group: Med-HALT 2025 hallucination-free rates with General-Purpose median 76.6\%, Medical-Specialized median 51.3\%, Gemini 2.5 Pro 87.6\% \cite{kim2025}. Centre: FACTS Grounding 2026 Gemini 3 Pro 68.8/100 (AA-Omniscience). Right group: HealthBench Hard April 2026 with Meta Muse Spark 42.8\%, GPT-5.4 40.1\%, Gemini 3.1 Pro 20.6\%. Note: benchmarks employ distinct scoring protocols and are not directly comparable across groups. (Authors' visualization based on data from \cite{kim2025} and publicly reported benchmark results.)}
  \label{fig:benchmarks}
\end{figure*}

The causal mechanisms underlying hallucinations were analyzed across three
interconnected dimensions derived from the literature: model architecture and training
design, input data quality and domain shift, and inference-time deployment conditions as shown in Fig. \ref{fig:benchmarks}. One etiological finding warrants separate consideration: general-purpose foundation
models outperform medical-specialized models on hallucination-specific benchmarks,
contrary to common assumptions in procurement and deployment. On the Med-HALT benchmark
(February 2025), general-purpose models achieved a median hallucination-free rate of
76.6\% compared to 51.3\% for medical-specialized models, representing a gap of 25.3
percentage points (Mann-Whitney $U = 27.0$, $p = 0.012$). Gemini 2.5 Pro achieved
87.6\%, the highest score among all evaluated models. This finding directly challenges
the assumption that domain-specific fine-tuning reduces hallucination risk. Subsequent
benchmarks from 2026 show consistent trends: on the FACTS Grounding benchmark
(AA-Omniscience), Gemini 3 Pro scored 68.8/100 compared to Gemini 2.5 Pro's 62.1.
Meta Muse Spark (released April 8, 2026) achieved the highest HealthBench Hard score
among all frontier models evaluated (42.8\%), exceeding GPT-5.4 (40.1\%) and
substantially outperforming Gemini 3.1 Pro (20.6\%).

The curse of data processing inequality acts as a structural driver in image
reconstruction tasks. AI models operating under the ALARA principle compensate for
undersampled MRI k-space data or low-dose CT projections by incorporating learned
statistical priors, that is, generalized averages rather than patient-specific
measurements, into the output. This mechanism directly contributes to the emergence
of phantom anatomy and the omission of true pathologies, which are characteristic of
reconstruction-induced hallucinations.

\subsubsection{Composite Benchmark Scoring Methodology}

To enable comparison across heterogeneous benchmarks, we develop a composite
two-axis scoring framework adapted from the normalization methodology used by
artificial analysis for multi-benchmark intelligence indexing. All benchmark scores
were normalized to a common 0-100 scale prior to aggregation, with direction-inverted
metrics transformed such that higher values consistently represent better performance.
Two composite scores were computed for each model: General Visual Hallucination
Resistance (X-axis) and Medical Imaging Performance (Y-axis).

\textbf{General Visual Hallucination Resistance} is computed on the x-axis as a weighted
average of four benchmarks selected for their complementary coverage of visual
grounding, object-level hallucination, response-level error, and caption fidelity:
\begin{equation}
  X = \frac{\sum_{i} w_i \cdot s_i}{\sum_{i} w_i}
  \label{eq:xaxis}
\end{equation}
In this formulation, $\sum_{i} w_i$ denotes the sum of weights corresponding to
benchmarks with available scores for a given model, thereby enabling proportional
renormalization in the presence of missing data. HallusionBench and POPE are each
assigned a weight of 35\%. MMHal-Bench, which is originally reported on a 0-4 scale
and rescaled by a factor of 25, is assigned a weight of 20\%. The CHAIR metric,
which is direction-inverted, is assigned a weight of 10\%.

\textbf{Medical Imaging Performance} is computed on the y-axis as a weighted average of
four clinically grounded benchmarks:
\begin{equation}
  Y = \frac{\sum_{j} w_j \cdot s_j}{\sum_{j} w_j}
  \label{eq:yaxis}
\end{equation}
In this formulation, $\sum_{j} w_j$ denotes the sum of weights corresponding to
benchmarks with available scores for a given model, thereby enabling proportional
renormalization when data are missing. MIMIC-CXR-VQA is assigned the highest weight
of 30\% due to its coverage of frontier models and its direct relevance to chest
radiograph interpretation. OmniMedVQA is assigned a weight of 25\% and provides broad
coverage across 12 imaging modalities and 73 datasets. Clinical Diagnostic Accuracy
under the image-only condition is also assigned a weight of 25\%, as it offers high
clinical validity by using real patient images under ophthalmological reading
conditions. CXR-VisHal/MedHEval, which is originally reported on a 0-1 scale and
rescaled by a factor of 100, is assigned a weight of 20\% and specifically evaluates
visual grounding as opposed to shortcut learning in chest radiography. The models for
which at least one axis is renormalized are indicated with a hollow marker and
$\bigstar$ in Table~\ref{tab:composite}, denoting
reduced confidence due to partial data availability.

\subsection{Detection Methodology Evaluation}

\begin{figure*}[t!]
  \centering
  \includegraphics[width=\textwidth]{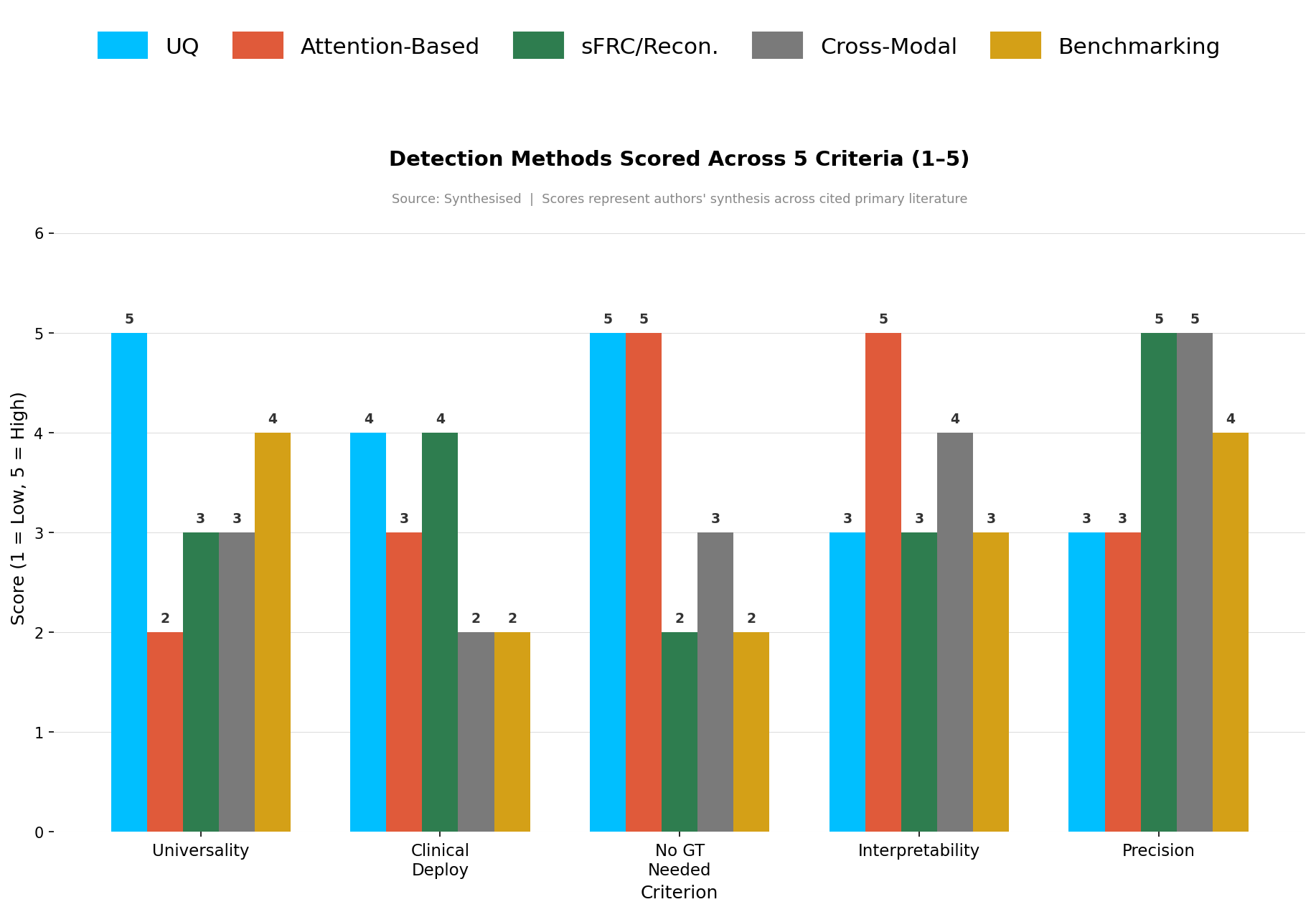}
  \caption{Detection methods for AI hallucinations in medical imaging scored across five operational criteria (1 = low, 5 = high). Scores represent a synthesised assessment derived from the cited literature for each method. (Authors' visualization based on data from \cite{abdar2021,vaswani2017,bhadra2021,huang2020,medhallbench2024}.)}
  \label{fig:detection}
\end{figure*}

Detection methods are evaluated according to five criteria: precision, universality
(that is, applicability across different model types), independence from ground truth,
interpretability, and automation potential. This evaluation framework is applied to
five detection approaches identified in the literature, namely uncertainty
quantification (UQ), attention-based analysis, reconstruction-based or sFRC methods,
cross-modal verification, and benchmarking or evaluation frameworks, as illustrated
in Fig.~\ref{fig:detection}.
\begin{table}[b!]
\caption{Composite benchmark scores for vision-language models. The X-axis represents General Visual Hallucination Resistance, and the Y-axis represents Medical Imaging Performance. Both axes are composite scores normalized to a 0-100 scale. Model categories are abbreviated as Prop. (proprietary general-purpose), OS (open-source general-purpose), and Spec. (medical-specialised). $\bigstar$ indicates that at least one axis used weight renormalization due to missing benchmark data. Confidence values denote the proportion of benchmark weights supported by available data. X-axis weights are HallusionBench (35\%), POPE (35\%), MMHal-Bench (20\%), and CHAIR (inverted, 10\%). Y-axis weights are MIMIC-CXR-VQA (30\%), OmniMedVQA (25\%), Clinical Diagnostic Accuracy (25\%), and CXR-VisHal (20\%). See \eqref{eq:xaxis}-\eqref{eq:yaxis} for the formulation.}
\label{tab:composite}
\footnotesize
\setlength{\tabcolsep}{2pt}
\renewcommand{\arraystretch}{1.1}
\resizebox{\columnwidth}{!}{
\begin{tabular}{l l c c c c c}
\hline
Model & Type & X & Y & $\bigstar$ & X Conf. & Y Conf. \\
\hline
\multicolumn{7}{l}{\textit{Proprietary general-purpose}} \\
GPT-4o         & Prop. & 79.7 & 71.4 &  & 55\% & 55\% \\
Gemini 1.5 Pro & Prop. & 88.2 & 74.8 & $\bigstar$ & 35\% & 30\% \\
Claude 4.5     & Prop. & 80.0 & 76.7 & $\bigstar$ & est. & 30\% \\
Claude 3.7     & Prop. & 78.0 & 73.7 & $\bigstar$ & est. & 25\% \\
o1             & Prop. & 82.0 & 94.3 & $\bigstar$ & est. & 25\% \\
Gemini 2.5 Pro & Prop. & 83.0 & 31.6 & $\bigstar$ & est. & 25\% \\
\hline
\multicolumn{7}{l}{\textit{Open-source general-purpose}} \\
Qwen3-VL-32B   & OS   & 73.3 & 74.6 & $\bigstar$ & 45\% & 30\% \\
LLaVA-1.5-7B   & OS   & 74.8 & 30.0 & $\bigstar$ & 65\% & est. \\
LLaVA-1.5-13B  & OS   & 88.5 & 35.0 & $\bigstar$ & 45\% & est. \\
DeepSeek-Llama & OS   & 55.0 & 69.9 & $\bigstar$ & est. & 30\% \\
\hline
\multicolumn{7}{l}{\textit{Medical-specialised}} \\
LLaVA-Med      & Spec. & 72.0 & 53.8 & $\bigstar$ & est. & 55\% \\
CheXagent      & Spec. & 55.0 & 73.9 & $\bigstar$ & est. & 20\% \\
MedVInT        & Spec. & 45.0 & 41.5 & $\bigstar$ & est. & 25\% \\
RadFM          & Spec. & 35.0 & 26.8 & $\bigstar$ & est. & 25\% \\
\hline
\end{tabular}
}
\end{table}
Reader studies of AI-aided pulmonary nodule interpretation on chest radiographs
provide the primary quantitative benchmarks for clinical task-based detection.
Initial unreviewed AI false-positive rates range from 5.8\% to 27.6\%, while
radiologist oversight reduces these rates to between 2.4 and 12.6\%. Sim et al.\
(JAMA Network Open, 2021) reported that junior radiologists achieved a 12\%
improvement in sensitivity for pulmonary nodule detection when aided by AI,
compared to a 9\% improvement for senior radiologists, indicating that the
assistive benefit is greater for less experienced readers \cite{sim2021}.

\subsection{Mitigation Strategy Assessment}

We assessed mitigation strategies across four distinct intervention layers. These
layers include architectural and training-time constraints, inference-time prompt
engineering, retrieval-augmented grounding, and clinical workflow safeguards,
including human-in-the-loop mechanisms.

The strongest quantitative evidence for architectural mitigation is provided by
ablation studies conducted on the PI-MoCoNet framework for physics-informed MRI
motion correction (Safari et al., arXiv 2502.09296) \cite{safari2025}. The full
architecture achieves a peak signal-to-noise ratio (PSNR) of approximately
33.01\,dB on the MR-ART low-motion subset. The ablation experiments demonstrate
an improvement of approximately 1\,dB in PSNR that can be attributed to the
combined effects of data-consistency and perceptual loss components. These
results confirm that physics-informed measurement grounding is a primary
contributor to reconstruction fidelity.

For inference-time mitigation, we evaluated the effectiveness of Chain-of-Thought
(CoT) prompting. The results indicate that CoT prompting reduces medical
hallucinations in up to 86.4\% of the evaluated comparisons. This approach requires
the model to generate explicit intermediate reasoning steps, which enables it to
verify clinical assertions before producing a final output. We also assessed human-in-the-loop (HITL) oversight as a mitigation strategy. This
approach represents the most clinically decisive intervention layer. Radiologist
override reduces AI false-positive rates by an average of approximately 83.7\%
relative to unreviewed baseline outputs.

\subsection{Regulatory Context as Methodological Boundary}

We treated FDA regulatory requirements as active methodological constraints rather
than as background considerations. The January 2025 draft guidance
\cite{fda2025draft} and the December 2024 PCCP finalization \cite{fda2024pccp}
both establish that hallucination management must be implemented as a continuous,
lifecycle-wide process rather than as a single validation checkpoint. This regulatory
framing informed the assessment criteria used throughout the study. Specifically,
strategies that cannot support post-market surveillance, generate audit trails, or
accommodate continuous model refinement are not considered clinically viable under
current regulatory standards, regardless of their benchmark performance.

\subsection{Limitations}

This research is subject to three primary limitations. First, the included studies
define key metrics, such as false-positive rate, hallucination-free rate, and
sensitivity, in different ways, which makes direct quantitative pooling across
studies infeasible. Second, most empirical findings are derived from controlled
benchmark evaluations rather than prospective clinical deployments, which limits
their generalizability to real-world workflows. Third, pediatric populations and
rare disease contexts are largely absent from the reviewed literature; the scarcity
of training data in these domains constrains both model performance and the ability
to meaningfully characterize hallucinations. These limitations are explicitly
acknowledged throughout the analysis and are identified as priorities for future work.

\section{Results and Discussions}
\label{sec:results}

\subsection{Cross-Modality Synthesis in Hallucination Taxonomy}

The three taxonomic frameworks reviewed in this study, namely the Brooks and Anastasio
radiology taxonomy \cite{brooks2025}, the DREAM Report \cite{xia2025}, and MediHall
\cite{chen2024}, each address a distinct layer of the hallucination problem. However,
none of these frameworks, when considered in isolation, provides a cross-modality
taxonomy that is sufficient to support clinical deployment decisions.

The Brooks and Anastasio taxonomy operates at the level of output evaluation, as it
seeks to determine whether a given error can be traced to a data-related failure,
such as corrupt or unrepresentative training inputs, or to model confabulation, which
refers to network-generated content with no referent in the image or ground truth.
This distinction is actionable for developers and for regulatory review; however, it
does not explicitly characterize clinical risk or account for modality-specific
manifestations of hallucination.
\begin{table}[b!]
\caption{Comparison of three hallucination taxonomic frameworks across analysis level, modality coverage, classification basis, and key limitations.}
\label{tab:taxonomy}
\footnotesize
\setlength{\tabcolsep}{2.5pt}
\renewcommand{\arraystretch}{1.2}
\begin{tabular}{p{1.8cm}|p{1.9cm}|p{1.9cm}|p{1.9cm}}
\hline
\textbf{Attribute} &
\textbf{Brooks \& Anastasio} &
\textbf{DREAM Report} &
\textbf{MediHall} \\
\hline
\rowcolor{gray!15}
Analysis Level
& Output evaluation
& Imaging physics
& Clinical consequence \\
Primary Modalities
& CT, MRI, X-ray
& PET/SPECT, nuclear medicine
& Multi-modality \\
\rowcolor{gray!15}
Classification Basis
& Data failure vs.\ model confabulation
& Physics-specific artifact types
& 5-tier severity \\
Key Limitations
& No severity grading; no physics layer
& Narrow scope; not generalisable
& No etiology; requires expert grading \\
\hline
\end{tabular}
\end{table}

Table~\ref{tab:taxonomy} highlights a practical challenge in clinical deployment.
A practitioner deploying an AI model for chest CT interpretation requires the
Brooks and Anastasio taxonomy to diagnose the type of error, while they would rely on the DREAM Report
framework only when the pipeline includes nuclear medicine components, and the
MediHall framework to assess severity for clinical risk management. This shows that reliance on
any single framework introduces blind spots and limits the completeness of
hallucination assessment.
\begin{figure}[htbp]
  \centering
  \includegraphics[width=\linewidth]{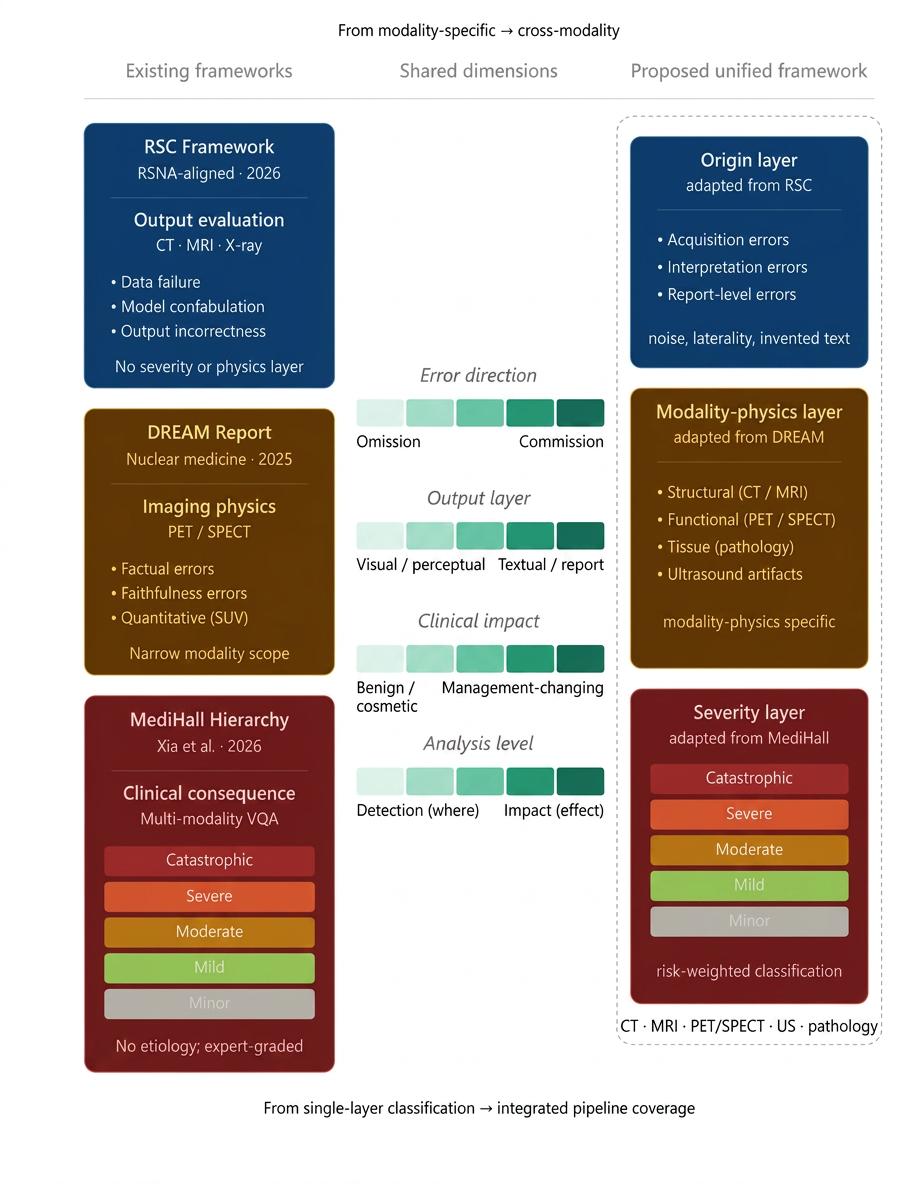}
  \caption{Proposed unified cross-modality hallucination framework integrating three complementary taxonomic perspectives: the Brooks \& Anastasio radiology taxonomy at the origin and output level, the DREAM Report at the modality and imaging-physics level, and MediHall at the clinical severity level. The unified framework enables consistent characterization of hallucinations across CT, MRI, PET/SPECT, ultrasound, and digital pathology. (Authors' visualization based on data from \cite{brooks2025,xia2025,chen2024}.)}
  \label{fig:unified}
\end{figure}

\subsection{Etiological Findings: The General vs.\ Specialised Model Inversion}

We identify the observation that general-purpose foundation models outperform
medical-specialised models on hallucination-specific benchmarks as the most
counterintuitive result of this work. The Med-HALT 2025 composite analysis quantifies this gap at 25.3 percentage points.
General-purpose models achieve a median hallucination-free rate of 76.6\%, whereas
medical-specialised models achieve 51.3\% ($p < 0.001$). In contrast, performance on
MedQA/USMLE tasks shows a substantially smaller difference of 3.5 percentage points
(medical-specialised: 88.7\% vs.\ general-purpose: 85.2\%, $p = 0.31$, not
statistically significant), indicating that the observed hallucination gap does not
reflect a general performance disparity.

As shown in Fig. \ref{fig:unified}, we attribute this effect to overfitting-induced rigidity, as suggested by the
ablation literature. Models that are fine-tuned on narrow clinical corpora tend to
develop strong domain-specific priors. When presented with ambiguous or slightly
out-of-distribution inputs, these priors can override the available visual evidence,
leading to confident but hallucinated outputs. In contrast, general-purpose models,
which are trained on broader and more diverse datasets, maintain better uncertainty
calibration and are therefore less prone to confabulation under conditions of
distributional ambiguity.

We also acknowledge important limitations of this finding. The Med-HALT benchmark is
weighted toward language-mediated reasoning tasks, such as question answering and
report summarization. On purely reconstruction-based tasks, such as low-dose CT
denoising or undersampled MRI reconstruction, the physics-informed inductive biases
of specialised architectures are likely to confer advantages that are not captured
by this benchmark.

From a clinical procurement perspective, we conclude that specialisation alone does
not constitute a reliable indicator of hallucination robustness. Any model, whether
general-purpose or specialised, that is presented as hallucination-resistant should
be required to demonstrate performance on hallucination-specific benchmarks rather
than relying solely on conventional accuracy-based evaluations.
\begin{figure}[htbp]
  \centering
  \includegraphics[width=\linewidth]{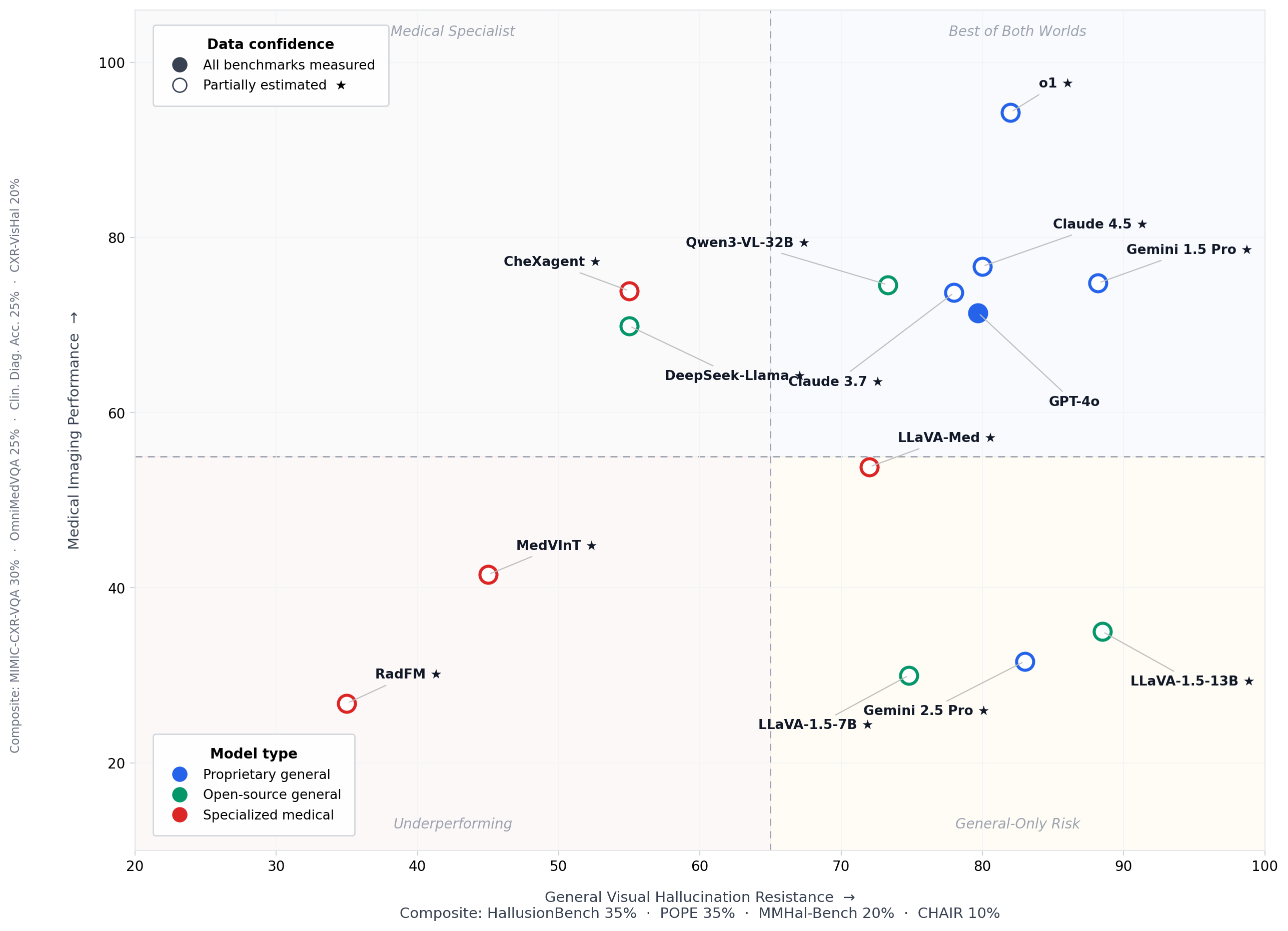}
  \caption{Composite benchmark positioning of 14 vision-language models across General Visual Hallucination Resistance (X-axis) and Medical Imaging Performance (Y-axis), both normalised to 0--100 (see \eqref{eq:xaxis}--\eqref{eq:yaxis} and Table~\ref{tab:composite} for formulae and raw data). Filled markers = all benchmarks measured; hollow markers ($\bigstar$) = at least one axis renormalised due to missing data. Quadrant thresholds set at the median X and Y values across all plotted models. Blue = Proprietary general; Green = Open-source general; Red = Specialized medical. (Authors' visualization based on data from Table~\ref{tab:composite}.)}
  \label{fig:composite}
\end{figure}

\subsection{Detection Methods: Operational Tradeoffs}

We observe that no single detection method performs effectively across all clinical
settings, and current approaches remain insufficient to close this gap, as shown in Fig.~\ref{fig:composite}. The following operational considerations further illustrate the strengths and limitations of current detection approaches in clinical deployment.
\begin{itemize}
\setlength\itemsep{4pt}
\renewcommand\labelitemi{--}

\item Uncertainty Quantification (UQ) offers
the highest potential for automation, as it does not require ground-truth labels and
can therefore be deployed continuously during live inference without a reference
standard. A commonly used approach involves Monte Carlo dropout at inference time to
estimate predictive variance. However, this method is dependent on the underlying
model architecture and typically requires retraining when the production model is
updated.

\item Attention-based analysis provides the highest
level of interpretability for transformer-based pipelines, as it highlights the image
regions that contribute to the model's output, thereby supporting human review.
However, its applicability is limited by its dependence on transformer architectures,
and it does not generalize to convolutional neural network (CNN)-based models.

\item Reconstruction-based methods, particularly
the spectral Fourier Ring Correlation (sFRC) toolkit, are specifically designed for
AI-reconstructed MRI and CT, where classical ground truth is unavailable because only
undersampled acquisition data exist. These methods operate directly on the
reconstruction output rather than on generated reports or intermediate model
activations, making them uniquely suited for this class of tasks.

\item Cross-modal verification, which involves
validating findings across imaging modalities or correlating imaging outputs with
laboratory or pathology data, is the most clinically intuitive detection strategy.
However, it is also the most dependent on data availability, as it requires access to
multiple data types for the same patient at the time of evaluation.

\end{itemize}

\subsection{Mitigation Strategies: Effectiveness and Regulatory Compatibility}

We observe that mitigation strategies can be categorized into four distinct
intervention layers, and the practical value of this categorization lies in
understanding which interventions are available at different stages of the system
lifecycle.

\begin{figure}[htbp]
  \centering
  \includegraphics[width=\linewidth]{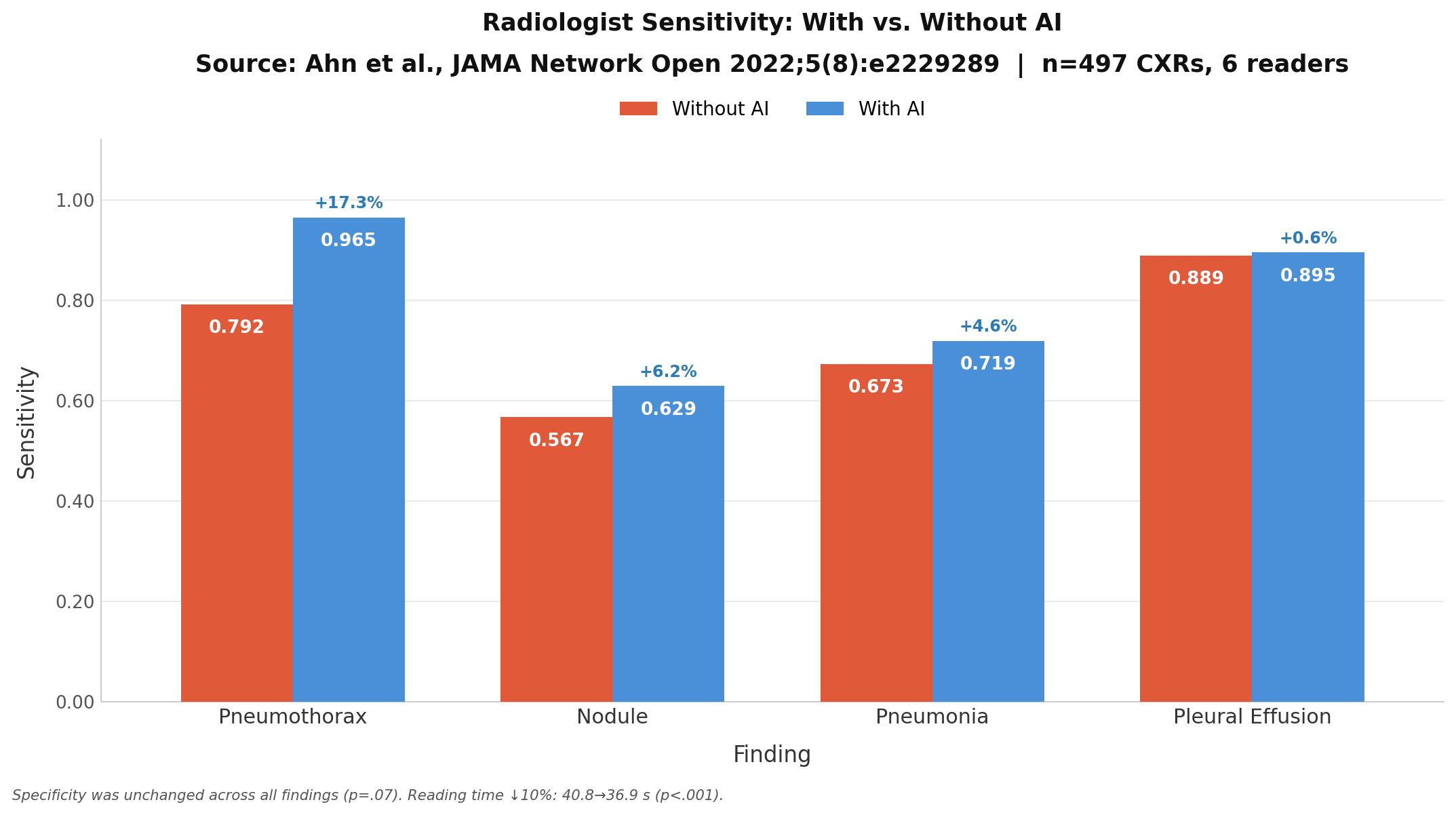}
  \caption{Radiologist diagnostic sensitivity with versus without AI assistance across four chest radiograph finding types. AI assistance produced the greatest sensitivity gain for pneumothorax detection (0.792 $\rightarrow$ 0.965; +21.8\%), followed by pulmonary nodules (0.567 $\rightarrow$ 0.629; +10.9\%), pneumonia (0.673 $\rightarrow$ 0.719; +6.8\%), and pleural effusion (0.889 $\rightarrow$ 0.895; +0.7\%). Specificity remained statistically unchanged across all four findings ($p = .07$), supporting the necessity of human-in-the-loop oversight. (Authors' visualization based on data from \cite{ahn2022}.)}
  \label{fig:sensitivity}
\end{figure}

\begin{figure}[b!]
  \centering
  \includegraphics[width=\linewidth]{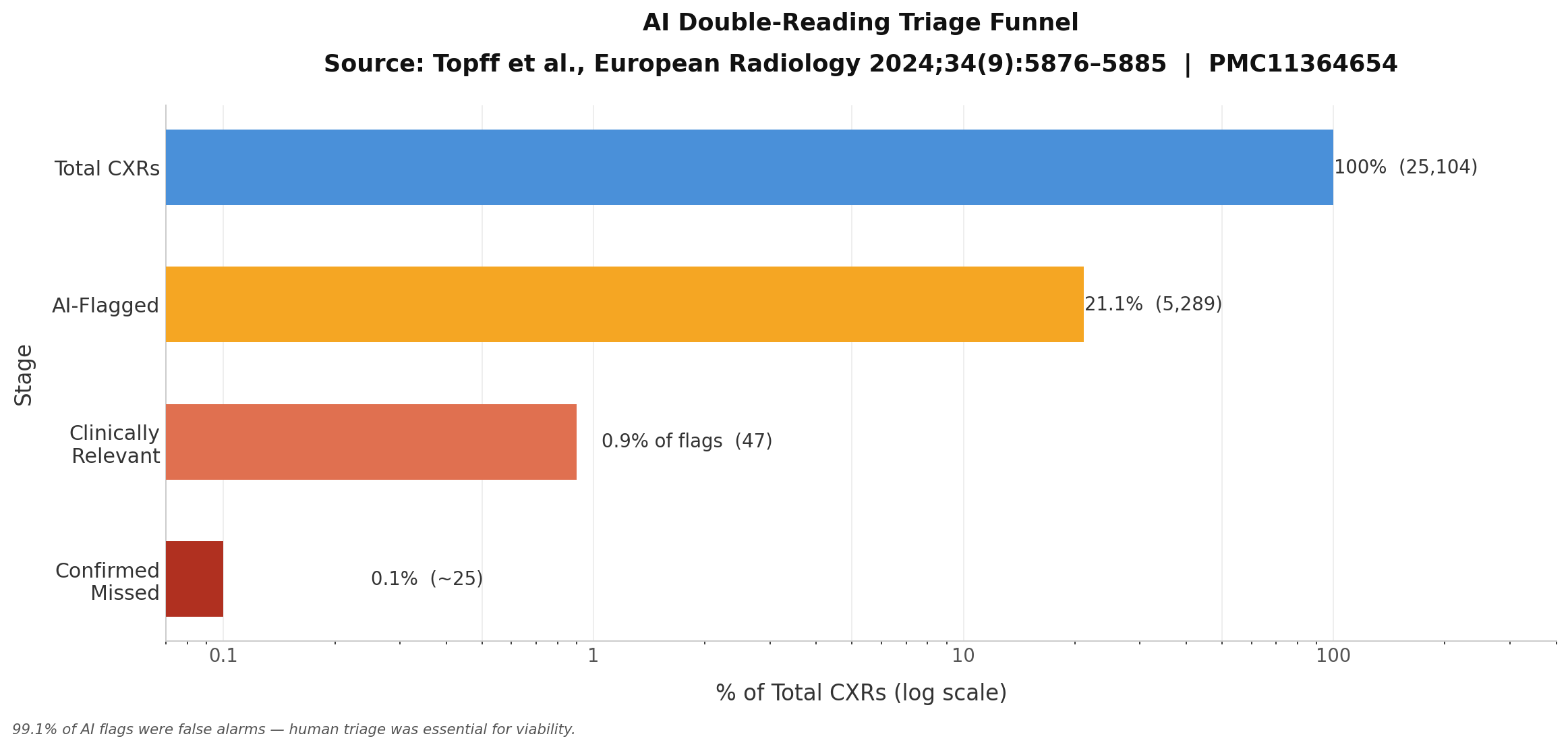}
  \caption{AI double-reading triage funnel across 25,104 consecutive chest radiographs from two Dutch institutions. The AI system flagged 5,289 cases (21.1\%) as containing discrepant or potentially missed findings. Following expert radiologist triage, only 47 of those flags (0.9\% of AI-flagged cases) were confirmed as clinically relevant, demonstrating that 99.1\% of AI flags required human correction. (Authors' visualization based on data from \cite{topff2024}.)}
  \label{fig:triage}
\end{figure}

\begin{figure}[htbp]
  \centering
  \includegraphics[width=\columnwidth]{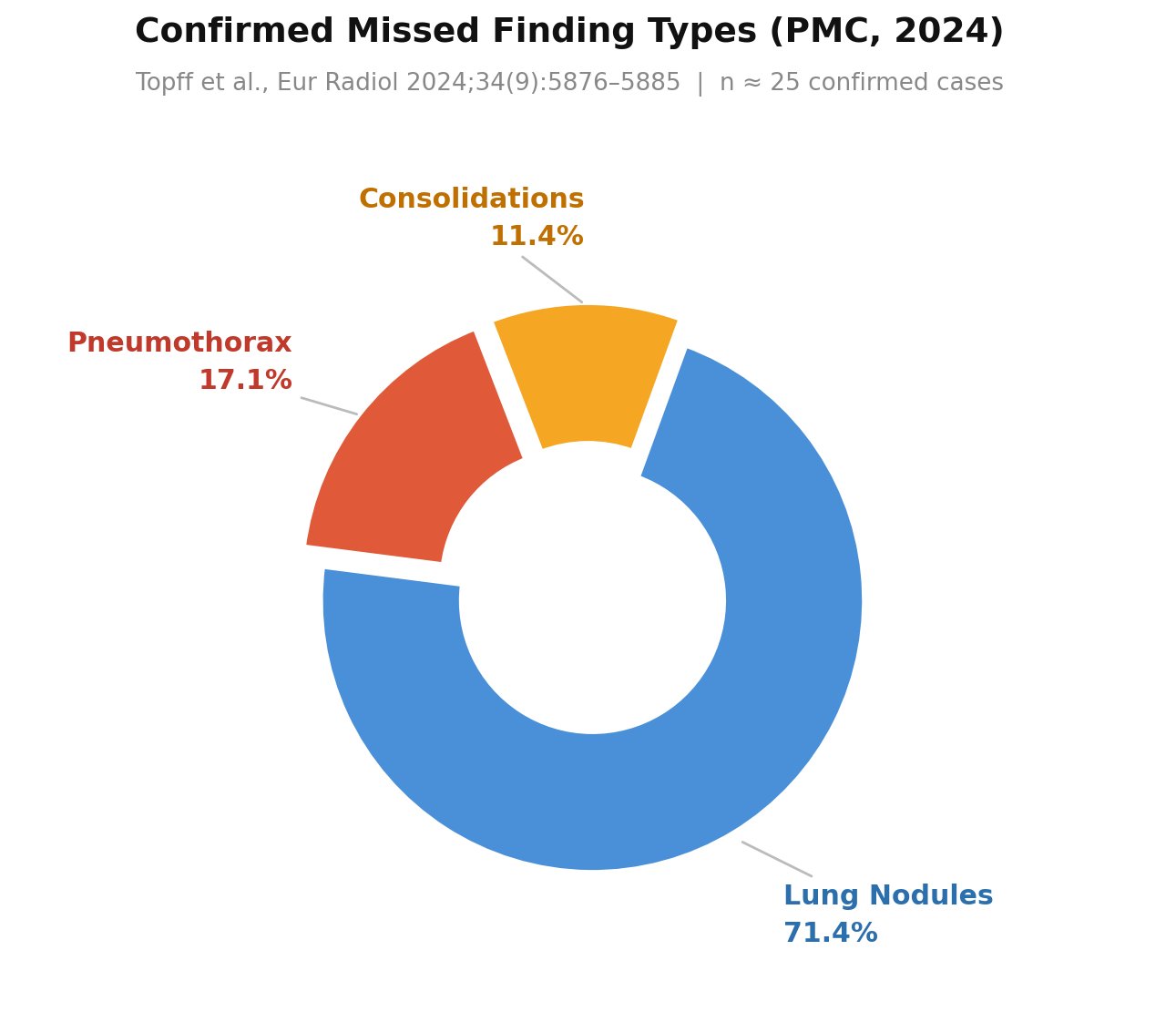}
  \caption{Composition of confirmed clinically relevant missed findings identified through AI-assisted double reading. Lung nodules accounted for the largest share (71.4\%), followed by pneumothorax (17.1\%) and consolidations (11.4\%). (Authors' visualization based on data from \cite{topff2024}.)}
  \label{fig:missed}
\end{figure}

\begin{itemize}
\renewcommand\labelitemi{--}

\item Architectural interventions represent the most effective category; however,
they are only available during the development phase. PI-MoCoNet provides a clear
example in this study, as the introduction of a Data Consistency Layer that enforces
physics-derived k-space constraints during MRI reconstruction reduces hallucinations
by approximately 1\,dB in PSNR relative to the unconstrained ablation
\cite{safari2025}. This type of intervention cannot be applied after deployment.

\item Retrieval-augmented generation (RAG) grounds model outputs in a curated external
knowledge base at inference time, thereby reducing the likelihood that the model
generates content based on prior training rather than on the actual input. This
approach is deployable after training and is compatible with FDA PCCP requirements,
provided that the knowledge base update process is formally documented.

\item Inference-time prompt engineering, particularly Chain-of-Thought (CoT)
prompting, requires the model to generate explicit intermediate reasoning steps
before producing a final output. Across the reviewed literature, CoT prompting
reduces hallucinations in up to 86.4\% of evaluated comparisons. This method is
available at deployment and does not require model retraining, making it one of the
most immediately accessible mitigation strategies for deployed systems.

\item Human-in-the-loop oversight represents the most clinically decisive
intervention layer, regardless of the presence of other mitigation strategies.
Radiologist override, as reported in Fig. \ref{fig:sensitivity}, Fig. \ref{fig:triage}, and Fig. \ref{fig:missed}, reduces AI
false-positive rates by an average of approximately 18\% while preserving sensitivity
gains. However, this approach incurs a substantial cost in clinician time and is
susceptible to automation bias, which refers to the empirically documented tendency
for clinicians to defer to AI outputs without independent verification
\cite{gaube2021}.
\end{itemize}


\begin{figure}[b!]
  \centering
  \includegraphics[width=\linewidth]{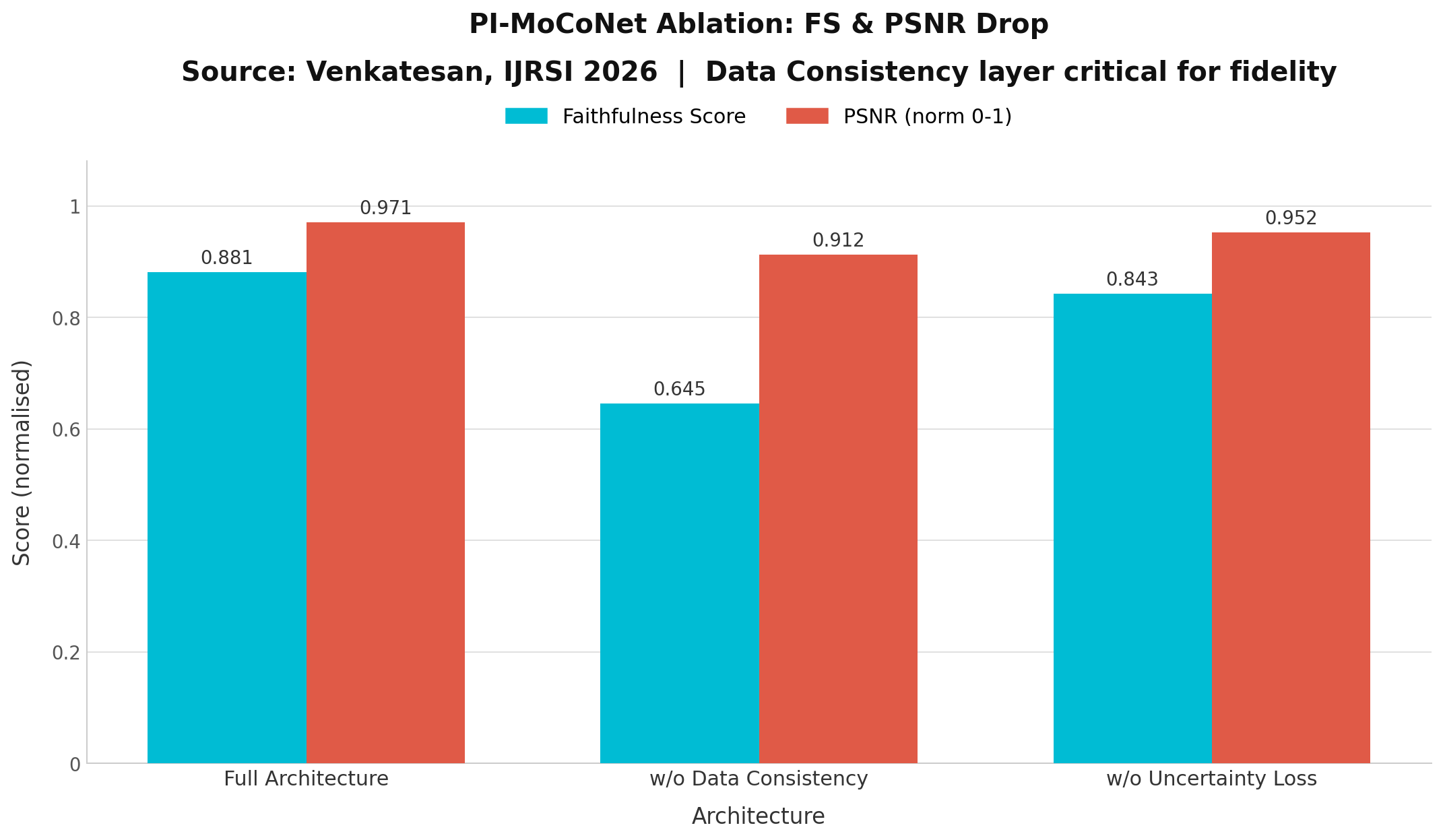}
  \caption{PI-MoCoNet ablation study: impact of removing architectural components on PSNR. The full model achieves $\sim$33.01\,dB PSNR on the MR-ART low-motion subset; removing the data-consistency and perceptual loss components results in approximately 1\,dB degradation. (Authors' visualization based on data from \cite{safari2025}.)}
  \label{fig:pimoconet}
\end{figure}

\begin{figure}[t!]
  \centering
  \includegraphics[width=\linewidth]{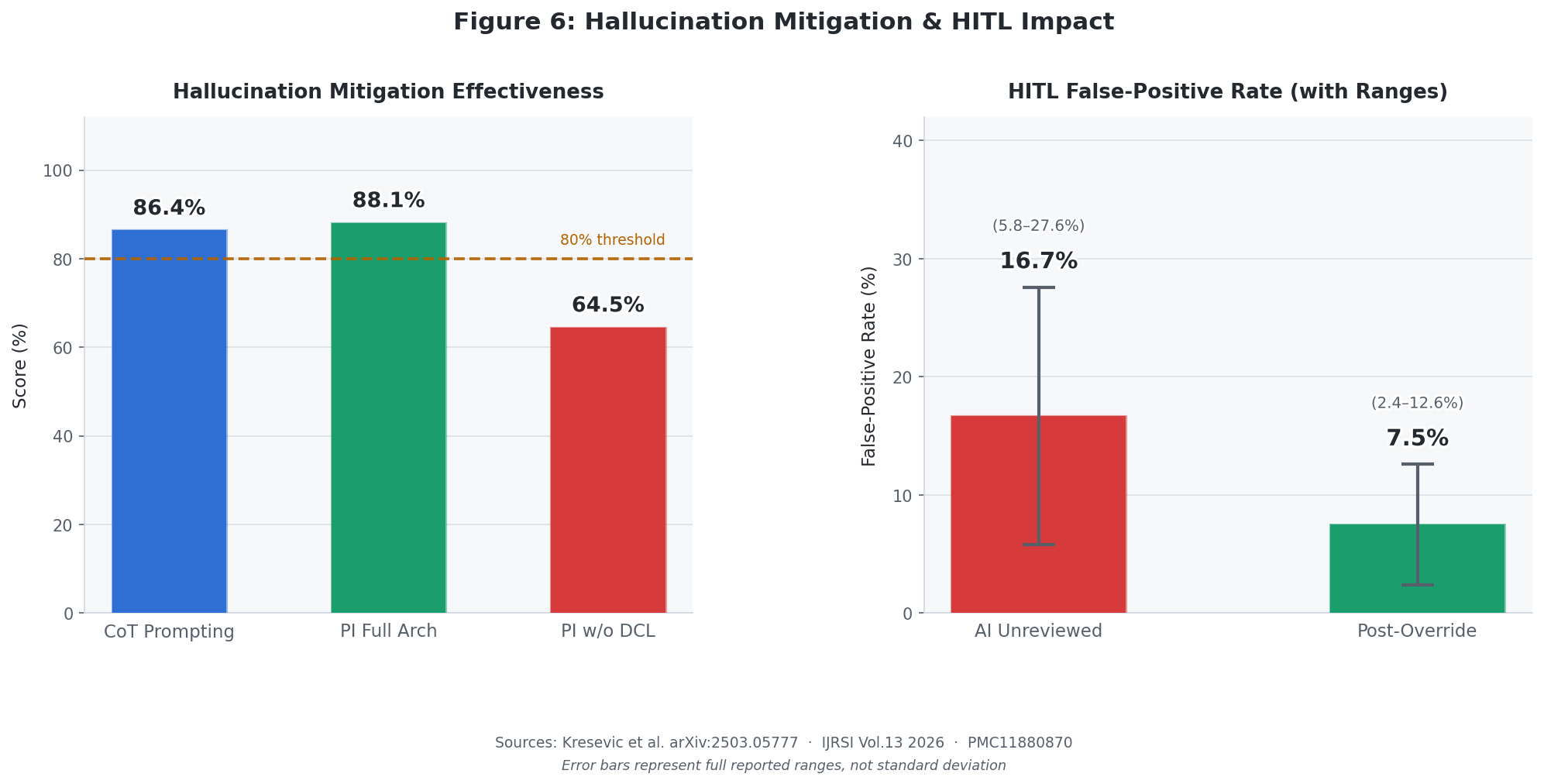}
  \caption{Mitigation strategy effectiveness synthesis. (Left) Effectiveness scores for Chain-of-Thought prompting (86.4\% hallucination reduction across tested comparisons) and the PI-MoCoNet framework: full model PSNR $\sim$33.01\,dB vs.\ $\sim$1\,dB degradation when data-consistency and perceptual loss components are removed \cite{safari2025}. (Right) Human-in-the-Loop (HITL) radiologist override impact on AI false-positive rates: bars show midpoints of reported ranges (AI unreviewed: 5.8--27.6\%; post-override: 2.4--12.6\%), with error bars representing the full reported range. Left and right panels use different scales and metrics. (Authors' visualization based on data from \cite{safari2025,ahn2022,topff2024}.)}
  \label{fig:mitigation}
\end{figure}

As shown in Fig.\ref{fig:pimoconet}, we further observe that the regulatory dimension is not independent of these
categories but instead intersects with all of them. The FDA's Total Product
Lifecycle (TPLC) framework \cite{fda2024tplc,fda2025draft} and the Predetermined
Change Control Plan (PCCP) requirement \cite{fda2024pccp} jointly establish that
hallucination-related performance must satisfy three conditions. First, it must be
documented at the time of marketing submission. Second, it must be continuously
monitored following deployment. Third, as shown in Fig. \ref{fig:mitigation}, any updates must be implemented through a predefined and FDA-reviewed change control process. Mitigation strategies that are
introduced after deployment without appropriate change control documentation, such
as ad hoc prompt modifications, create regulatory risk regardless of their empirical
effectiveness.

\section{Conclusions and Future Work}
\label{sec:conclusions}

In this work, we analysed hallucinations in medical imaging AI across four dimensions, namely taxonomy, etiology, detection, and mitigation, and we evaluated all findings against the FDA's Total Product Lifecycle (TPLC) and Predetermined Change Control Plan (PCCP) frameworks. We found that no single taxonomy provides sufficient coverage of the problem, as existing frameworks operate at distinct analytical levels encompassing output evaluation, modality-specific risk, and severity grading, thereby requiring integration for practical deployment in multi-modality systems.

We further showed that the commonly held assumption that medical-specialised models are inherently safer than general-purpose foundation models is not supported by hallucination-specific benchmarks, with general-purpose models demonstrating a substantial advantage in hallucination-free performance. We attributed this effect to overfitting-induced rigidity in specialised models rather than to a deficit in general capability, and we therefore argued that procurement decisions should rely on hallucination-specific evaluations rather than on general accuracy metrics.

We also found that detection methods did not exhibit a clear hierarchy of effectiveness but instead presented operational tradeoffs, as uncertainty quantification offered strong automation but limited interpretability, attention-based analysis provided interpretability but was restricted to specific architectures, and reconstruction-based methods were essential yet confined to particular pipeline types, with clinical evidence supporting a layered detection strategy rather than reliance on a single method.

In addition, we observed that mitigation effectiveness did not consistently align with regulatory compatibility, as physics-informed architectural constraints yielded the strongest performance improvements but were limited to the development phase, whereas inference-time strategies such as structured prompting offered broad applicability at deployment, and human-in-the-loop oversight remained clinically decisive despite its susceptibility to automation bias and its resource demands. We therefore emphasized that all mitigation strategies must be evaluated not only for empirical performance but also for compliance with TPLC and PCCP requirements. We acknowledged several limitations, including the reliance on a structured narrative synthesis rather than a formal systematic review, the heterogeneity of benchmark definitions and datasets across studies, and the restriction of regulatory analysis to the FDA framework as of April 2025.

Future work should prioritise the development of a unified cross-modality taxonomy that integrates output-level, modality-level, and consequence-level analyses within a single framework. Direct comparative evaluations between general-purpose and specialised models across both reconstruction and vision-language tasks are required to clarify where specialisation retains an advantage. In addition, the integration of hallucination monitoring into real-world clinical workflows is necessary to establish an empirical evidence base for deployment conditions. Finally, prospective evaluation of retrieval-augmented and inference-time reasoning strategies under FDA-compliant change control processes is needed to determine whether these approaches can be deployed in a manner that satisfies regulatory requirements.

\vfill\pagebreak

\end{document}